\documentclass[conference]{IEEEtran}
\IEEEoverridecommandlockouts
\usepackage{cite}
\usepackage{amsmath,amssymb,amsfonts}
\usepackage{graphicx}
\usepackage{textcomp}
\usepackage{xcolor}

\usepackage{filecontents}
\usepackage{multirow}
\usepackage{array, tabularx,makecell}
\usepackage{algorithm}
\usepackage{algpseudocode}
\usepackage{mathtools,nccmath}
\usepackage{lipsum,adjustbox}
\usepackage{pgfplots}
\usepackage{pgfplotstable}
\usepackage{siunitx}
\usepackage{subcaption}
\usepackage{caption}
\usepackage{pgfplots}
\newcommand{\argmin}{\arg\!\min}

\def\BibTeX{{\rm B\kern-.05em{\sc i\kern-.025em b}\kern-.08em
    T\kern-.1667em\lower.7ex\hbox{E}\kern-.125emX}}
    \newcommand{\R}{\mathbb{R}}
\begin{document}

\title{Context-Aware Design of Cyber-Physical Human Systems\ (CPHS)\
{
\thanks{This research was supported by Transportation Consortium of South-Central States (Tran-SET) Award No 18ITSLSU09/69A3551747016. Any opinions, findings, and conclusions or recommendations expressed in this material are those of the author(s) and do not necessarily reflect the views of the sponsor. }
}
}

\author{\IEEEauthorblockN{Supratik Mukhopadhyay\IEEEauthorrefmark{1}, Qun Liu\IEEEauthorrefmark{1}, Edward Collier\IEEEauthorrefmark{1}, Yimin Zhu\IEEEauthorrefmark{1},
Ravindra Gudishala\IEEEauthorrefmark{1}, \\Chanachok Chokwitthaya\IEEEauthorrefmark{1}, Robert DiBiano\IEEEauthorrefmark{3}, Alimire Nabijiang\IEEEauthorrefmark{1}, Sanaz Saeidi\IEEEauthorrefmark{1}, Subhajit Sidhanta\IEEEauthorrefmark{2}, 
Arnab Ganguly\IEEEauthorrefmark{1}
}

\IEEEauthorblockA{\IEEEauthorrefmark{1}Louisiana State University, Baton Rouge, LA, USA\\
}

\IEEEauthorblockA{\IEEEauthorrefmark{2}Indian Institute of Technology, Bhilai, India
}
\IEEEauthorblockA{
\IEEEauthorrefmark{3}Ailectric LLC\\
supratik@csc.lsu.edu
}
}

\maketitle

\begin{abstract}
Recently, it has been widely accepted by the research community that  interactions  between humans and cyber-physical infrastructures have played a significant role in determining the performance of  the latter.

The existing paradigm for designing cyber-physical systems for optimal performance focuses on developing models based on historical data. The impacts of context factors driving human system interaction are challenging and are difficult to capture and replicate in existing design models. As a result, many existing models do not or only partially address those context factors of a new design owing to the lack of capabilities to capture the context factors. This limitation in many existing models often causes performance gaps between predicted and measured results.  
We envision a new design environment, a cyber-physical human system (CPHS) where decision-making processes for physical infrastructures under design are intelligently connected to distributed resources over cyberinfrastructure such as experiments on design features and empirical evidence from operations of existing instances. The framework combines  existing design models with context-aware design-specific data involving human-infrastructure interactions in new designs, using a machine learning approach to create  augmented design models with improved predictive powers. 
\end{abstract}


\section{Introduction}
Design (including engineering) is a key component for creating a virtuous cycle between existing and future infrastructure systems. Cyber-physical Infrastructure designs define their characteristics and functions according to goals and contexts of a  project. Cyber-physical infrastructure performance is an important factor during design that needs significant attention of  designers and engineers. Recently, it has been widely accepted by the research community that  interactions  between humans and cyber-physical infrastructures have played a significant role in determining the performance of  the latter \cite{nicol2004stochastic}. For example, while designing an energy efficient building, one has to take into account the behavior of the human users \cite{santin2009effect}.  However, human behavior depends on many context-dependent factors such as ethnic origin, gender, social status, political alignment, social connections, type of house, marital status, educational background, financial condition, etc \cite{o2014contextual}. Occupants’ interactions with building systems such as heating, ventilation, and air conditioning (HVAC), lighting, blinds, windows, and electronic appliances depend on various context factors \cite{kinateder2014social}. Since each building or occupant is unique, designing buildings, by optimizing performance with respect to human-building interactions, seems paradoxical.  However, improved methods for building design, optimizing performance, bears significant social, economic, and environmental consequences. The existing paradigm of designing buildings for optimal performance focuses on developing models based on historical data. The impacts of context factors driving human building interaction are challenging and are difficult to capture and replicate in design models. As a result, many existing models do not or only partially address those context factors of a new design owing to the lack of capabilities to model  the context factors accurately \cite{nicol2001characterising}. This limitation in many existing models often causes performance gaps between predicted and measured results.  

Similar is the case for transportation networks. Traffic management models that include routing choice \cite{ben2004route} form the basis of traffic management systems. These  models provide crucial  inputs towards predicting  traffic volumes on different routes and hence inform government policies for design and construction of new transportation artifacts as well as for designing  effective traffic management mechanisms that ensure minimum  traffic delays and maximum  usage of existing transport systems. High Fidelity models that are based on rapidly evolving contextual conditions can have a significant impact on smart and energy efficient transportation. Currently, existing traffic/route choice models are generic and are calibrated on static contextual conditions \cite{endsley1988situation}. These models do not consider evolving contextual factors such as the location, failure of certain portions of the road network,  the social network structure of population inhabiting the region (socio-cultural and economic background), route choices made by other drivers, events, extreme conditions, etc. As a result, the model’s predictions are made at an aggregate level and for a predetermined set of contextual conditions.  One needs to develop higher fidelity models wherein subjective and contextual factors are captured \cite{lima2016understanding} for making effective design and management decisions   for transportation systems. 
\subsection{Context-Aware Design: The  Case for Future Data}
Imagine you are the head of the State of Louisiana,  Department of Transportation. You are considering the creation of a light rail system between French Quarter, New Orleans  and the Garden district to reduce the average  travel time between these two places. The project will cost more than a hundred  million dollars. The money will be well-spent and the project will be successful if there is a considerable user-base for the system and it reduces the average travel time. How do you predict if people will use the system or whether it will reduce the average travel time? Of course, the only way of reliable prediction is based on data. How do you collect such data? The light-rail system does not even exist so that you can let people use it to collect data. Creating even a rudimentary system to collect data will cost millions of dollars (that can possibly be wasted if the system doesn't find enough usage). Of course, there are similar systems elsewhere in the country (Norfolk, VA,  Philadelphia, PA,  etc.). Data can certainly be collected from these. However, ridership depends on context; in general any phenomenon where human behavior is involved depends on context. The context in New Orleans may be different from those in places like Norfolk, Philadelphia and others. As a result, predictions based on data collected at other places may not closely reflect the ground truth at New Orleans. 

The above scenario happens in many endeavors. Consider the case of  designing  an energy efficient building as discussed above. You have to take into account the behavior of the human users. Or suppose  that you are designing the disaster evacuation policy for your city \cite{kobes2010exit}. In both these cases, data about human interaction with the system under design will only be available in the future after the system has been implemented. Also in both cases, the design of the system needs to take into account the human system interaction that affects the performance. 

Traditional AI techniques tend towards  an  ``average'' that minimizes a loss function \cite{bishop2006pattern}. However, human behavior depends on many context-dependent factors such as race, gender, social status, political alignment, social connections, marital status, educational background, financial condition,  etc. Both energy usage and evacuation, as mentioned above, depend strongly on human behavior. So an AI-based decision that ignores context (as for example a special occupational background) will be a poor predictor. Similarly, traditional game theory considers agents to be perfectly rational and therefore are not able to predict accurately in context dependent situations. However, collecting data on how these context dependent factors influence human behavior requires creating the building in the former case while in the latter case it requires creating  disasters; both extremely expensive. 

Traditional AI has in the recent years focused on Bigdata.  The assumption is that a large volume of labeled data is available. But in the  situations  shown above, acquiring such data at design time may be expensive and infeasible. 
\begin{figure}[t!]
    \centering
    \includegraphics[width=0.5\textwidth]{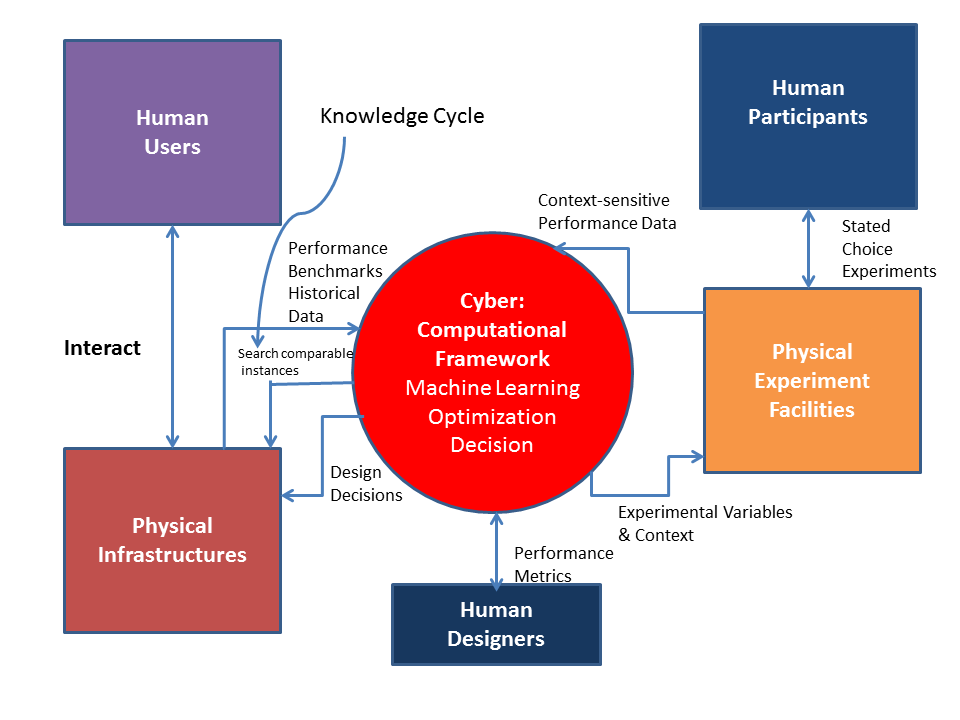}
    \caption{A CPHS to Support Human-Centered Design}
    \label{comp_result1}
\end{figure}
\subsection{The CPHS Framework}
We envision a new design environment, a cyber-physical human system (CPHS) where decision-making processes for physical infrastructures under design are intelligently connected to distributed resources over cyberinfrastructure such as experiments on design features and empirical evidence from operations of existing instances. With such a system, design for  performance (such as energy efficiency in buildings) is no longer an isolated process heavily dependent upon designer’s and engineer’s experience and interpretation of how humans possibly interact with systems under design. Rather, design becomes an inclusive process wherein designers and engineers can dynamically sample existing infrastructures for performance benchmarks and test their design features using collaborative experiment facilities (Fig. \ref{comp_result1}). Therefore, the CPHS involves different types of human, such as designers  who set performance criteria for infrastructures under design and make decisions, users of existing infrastructures of the same type where empirical operational data and  performance benchmarks are obtained, and participants of stated choice experiments from whom designers and engineers can better understand how design features impact human infrastructure interactions in different contexts; the physical components of the CPHS include experiment facilities and existing infrastructures; and the cyber component is a computational framework to integrate empirical data (from existing infrastructures) and experimental data (from experiments) in order to enhance performance predictions and support decision-making during design.  A feedback loop exists among those components. First, the designers and engineers set design criteria and initial performance metrics. Then, the computation framework  suggests comparable infrastructures to search for. With the approval of designers and engineers, initial performance benchmarks and operational data can be derived. With input from designers and engineers, a list of experiment variables and associated contextual conditions for experiments are suggested by the computational framework. After experiments, data specific to design contexts are used by the computational framework to optimize predictions of  performance. After examining the predictions, designers and engineers can adjust the performance metrics and design criteria to repeat the entire process, or a partial process (i.e., between the cyber component and the experiment facilities or the cyber component and existing infrastructure).

In essence,  our framework extends existing cyber-physical-human system models by including “future” data to support prediction and decision-making. This capability is essential to design, although the framework uses performance design as a focal point.  A traditional CPHS deals with actual physical component to provide feedback to decision-making or optimize its operations. In design, we often deal with conditions where the physical component does not exist. Instead, experiment facilities where virtual proxies of physical systems under design are used to acquire targeted data (which serve as proxies for the future data) to improve predictive models developed based on data from existing systems. Therefore, new AI techniques are needed to combine data from the experiment cycle and the knowledge cycle. Human interactions with selected design features are studied to understand its potential to influence design decision-making. 
\section{Mathematical Underpinnings of CPHS}
A CPHS design  is a set of design  (random) variables $\{X_1, \ldots, X_n\}$ assuming values in  respective domains $D_1, \ldots, D_n$, and a set of probability distributions $\pi_1, \ldots, \pi_n$, such that, for $1 \leq i \leq n$, $X_i \sim \pi_i$, where the support of $\pi_i$ is $D_i$. Let $\mathit{Des}$ be the class of CPHS designs. A CPHS design specification is a mapping $s: \mathit{Des} \rightarrow \R$. In a buidling design, for example, such a mapping can indicate energy savings for a particular design. The CPHS design problem is: for $d \in \mathit{Des}$  with set of design  (random) variables $\{X_1, \ldots, X_n\}$ assuming values in  respective domains $D_1, \ldots, D_n$, such that $s(d) = c \in \R$ and a (historical) data set $\mathit{Dat}$, estimate $d$.  

To understand the mathematical underpinnings behind the computational framework underlying future data CPHS, we first start with the theoretical foundations of existing machine learning paradigms.

\subsection{The Two Existing Paradigms of Machine Learning}

Consider a classification problem. Suppose that you have a random variable  $X$  that takes values over an an infinite set $\cal{X}$. Let $Y$ be another random variable that takes values over the set \{0, 1\}. We assume that the probability distribution $\mu$  that  $X$  follows is known (for example, it can be estimated through a survey).   Of course,  there exists a  target joint probability distribution of $(X,Y)$.  We are asked to design a function $g:\cal{X}$ $\rightarrow \{0, 1\}$. The test error produced by the function is given by $\mathit{err} = P(g(X) \neq Y)$ ($P$ represents  the unknown target probability) where $(X,Y)$ follows the  target joint distribution of $(X,Y)$ \cite{devroye2013probabilistic}. The goal of machine learning is to synthesize a function $g^*$ that minimizes this error. Mathematically, $g^* = \argmin_{g \in 2^{\cal{X}}}{P(g(X) \neq Y)}$ \cite{devroye2013probabilistic}. Traditional machine learning considers two cases.

\subsection{Bayesian Case: Known Joint Distribution}
The test error $P(g(X) \neq Y)$ depends on  the target  distribution on $(X,Y)$. Let us first assume that this \emph{target distribution is known}. In this case, we call the  test error, Bayesian error \cite{devroye2013probabilistic}. Let $\eta(x) = P(Y=1 \mid X=x)$ \cite{devroye2013probabilistic} where $x \in {\cal X}$.  Since the target distribution on $(X,Y)$ is known, the function $\eta$ is known as well.  In this case, the test error for a function $g \in 2^{{\cal X}}$  can be calculated as $1-E((I_{g(X)=1}\eta(X) + I_{g(X)=0}(1-\eta(X)))$ \cite{devroye2013probabilistic} where $I_{g(x)=1}$ is the indicator function for the set $\{x \in {\cal X} \mid g(x) =1\}$ and $E$ is expectation \cite{devroye2013probabilistic}. In Bayesian machine learning, one determines the optimal $g^*$ by minimizing this expression, i.e., by minimizing the Bayesian error.  Rather than searching over the infinite set $2^{{\cal X}}$ for an optimal $g^*$,   one can show that the following $g^*$ minimizes the Bayesian error \cite{devroye2013probabilistic}.
\[\begin{array}{lll}
g^*(x) =1  & if & \eta(x) >1/2 \\
~~~~~~=0 & otherwise & \\
\end{array} \] \cite{devroye2013probabilistic}.
\subsection{Unknown Joint Distribution}
The second paradigm of traditional AI considers the case where the target distribution on $(X,Y)$ is unknown; but one can efficiently draw random samples from this distribution. In this case, one draws $N$ random iid samples from this distribution. Let $S$ be the set of samples drawn. We call this set  $S$ the training dataset. Since the target distribution on $(X,Y)$ is unknown, $\eta$ is unknown as well. So we cannot determine the error. However, we can determine the error restricted to the training data. For a function $g \in 2^{{\cal X}}$, the training error is given by $\mathit{trerr}=|\{x \in S \mid g(x)\neq y\}|/|S|$.  Now Hoeffding's inequality states  that for any $\epsilon>0$,  $P(|\mathit{err}-\mathit{trerr}|>\epsilon )\leq  2\mathit{exp}(-2N\epsilon^2)$. So if the size of the training data is sufficiently large, the training error closely tracks the test error \cite{bishop2006pattern}.  One can determine $g^*$ in this case by minimizing the training error for a sufficiently large training dataset \cite{bishop2006pattern}. Thus $g^* = \mathit{argmin}_{g \in 2^{\cal X}} |\{x \in S \mid g(x)\neq y\}|/|S|$. This serves as the foundation for supervised learning \cite{bishop2006pattern}. 

\subsection{The New Machine Learning Paradigm  for ``Future Data"}
As one can see, none of the two existing paradigms cover the scenarios described in the Introduction. In the Bayesian case, we assume that the target distribution on $(X,Y)$ is known. In the second case, we assume that the target distribution on $(X,Y)$ is unknown but one can efficiently draw random samples from it.  The test error depends on the size of the training dataset. In the scenarios described in the Introduction, the artifact about which we are predicting does not even exist. Thus neither it is possible to know the target distribution nor it is possible to efficiently draw random samples from that distribution. 
\def\Omega{\pi}
Now we try to formalize  the  new paradigm. Assume that we have a specification distribution $\Omega$ on $X \times Y$ (specified by a specification function $s$) for which we know the mean $\mu$ and the variance $\sigma^2$. We are not allowed to draw samples from  $\Omega$ since that is either impossible or prohibitively expensive.  While we may not be able to draw random samples from $\Omega$, it may be possible to draw random samples from unknown distributions that are close to $\Omega$. More formally, let $\pi_1$, \ldots, $\pi_k$ be $k$ unknown  \emph{auxiliary} probability distributions over $(X,Y)$ that are known to be within a distance $\alpha$ from $\Omega$ (these distributions can arise from empirical operational data or experimental data from stated choice experiments). Here $\alpha$ is based on a suitable distance metric on probability distributions such as Wasserstein distance, Prokhorov-Levy distance, etc. Suppose that it is possible to efficiently draw random samples from each $\pi_i$ where $1 \leq i \leq k$ (based on virtual models of the original artifact, for example). 
Let ${\cal H} \subseteq Y^X$ be a hypothesis set of bounded continuous functions. For a distribution $f$ define the  hypothesis loss function \[{\cal L}(f, \Omega) = \mathit{sup}_{h \in {\cal H}} |E_f(h(x))-E_{\Omega}(h(x))|.\]
From \cite{Dudley2018real}, it is known that ${\cal L}(f,\Omega) =0$ iff $\Omega =f$. We assume that the hypothesis loss function is  efficiently computable. 
For simplicity, if we assume $Y=X$ and ${\cal H} = \{\mathit{id}\}$, where $\mathit{id}$ is the identity function,  the loss function ${\cal L}(f,\Omega) = |\mu_f-\mu|$. 

The following is an approximate  CPHS design problem in this new machine learning paradigm:
\begin{enumerate}
\item Let ${\cal D}$ be the class of all probability distributions over $X\times Y$. Given $k$ auxiliary distributions $f_1, \ldots, f_k \in {\cal D}$, that are within a distance $\alpha$ from a specification distribution $\Omega \in {\cal D}$ with mean $\mu$ and variance $\sigma^2$, with samples available only from the auxiliary distributions,  generate a distribution $\hat{\Omega}$ such that ${\cal L}(\hat{\Omega},\Omega)<\epsilon$ and the distance between $\hat{\Omega}$ and $\Omega$ is bounded by a parameter $\beta$.  
\end{enumerate}

\subsection{Learning from Contexts}
Contexts can be described by predicates. Suppose we know that the test data will come from a particular context. For example, in case of the light rail system of New Orleans, the possible users have a particular context. Let the predicate $\varphi(x)$ denote that context on the set ${\cal X}$. The  context-sensitive test error for $g \in 2^{{\cal X}}$ can then be defined by $\mathit{teerrcon}= P(g(X) \neq Y \mid (X,Y)_{{\cal X}_{\varphi}})$ where we abuse notation to denote by $(X,Y)_{{\cal X}_{\varphi}}$ to denote the restriction of $X$ to vary on the set ${\cal X}_{\varphi} = \{x \in {\cal X} \mid \varphi(x)\}$. Suppose that it is easy to draw random samples from $(X,Y)$ but it is hard to extract samples from the probability distribution $(X,Y)_{{\cal X}_{\varphi}}$. How can we synthesize a function $g^*$ that minimizes the context-sensitive test error? One idea would be to approximate $\varphi$ by   a predicate $\varphi'$ such that it is easy to draw random samples from $(X,Y)_{{\cal X}_{\varphi'}}$. Such a predicate can for example be created in a virtual environment. 
\section{A Practical Framework for Context-Aware Design of CPHS}
\begin{figure}[t!]
    \centering
    \includegraphics[width=0.5\textwidth]{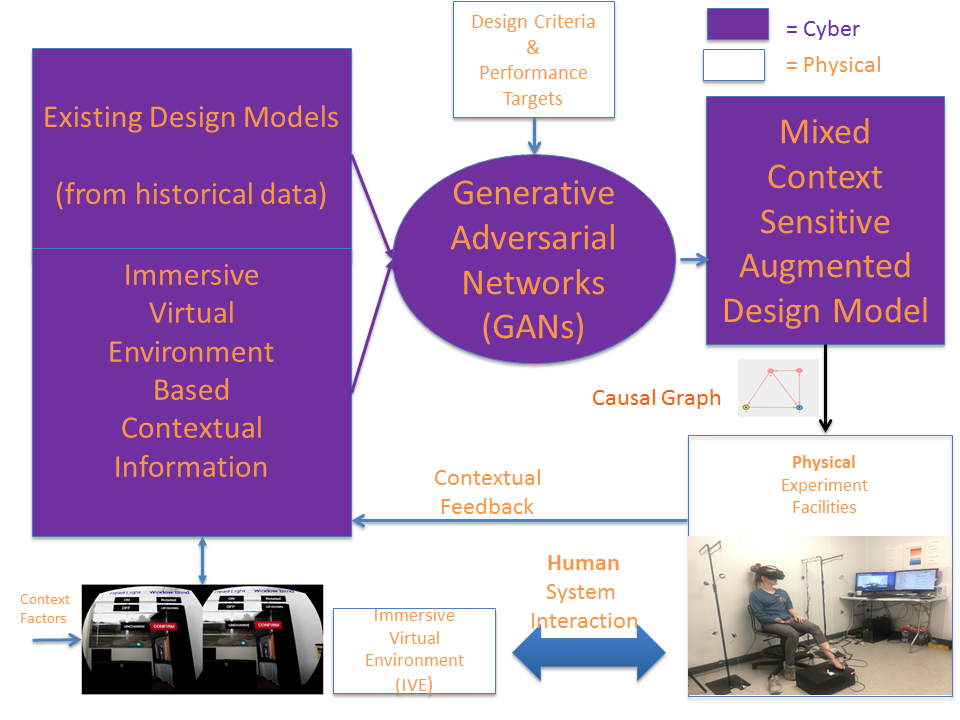}
    \caption{Architecture of the CPHS Framework}
    \label{comp_result2}
\end{figure}
Fig \ref{comp_result2} shows the architecture of our CPHS framework. Owing to the lack of ability to accurately model human interaction  in existing design models for new designs, we present the CPHS framework to enhance the predictive power of designs by appropriately combining existing designs (obtained from historical performance data) with contextual knowledge of human interactions  in a new design  obtained from IVE (Immersive Virtual Reality Environment) experiments \cite{chokwitthaya2017enhancing,chokwitthaya2019combining,chokwitthaya2019machine}. Generative Adversarial Networks (GANs) \cite{goodfellow2014generative} are used to generate  mixture models, that allow appropriate fusion of existing designs and knowledge of human interactions as obtained from stated choice  IVE experiments, with performance targets as guides. Experiments are conducted based on the mixture model obtained in physical experiment facilities and the feedback obtained is used for narrowing down/expanding/modifying the set of highly specific influent contextual factors that should be considered to improve predictions of the performance of the system under design in the next iteration.
\subsection{Identifying the contextual factors in human system  interactions}
Context-aware design-specific data capture contextual factors \cite{chokwitthaya2019combining} affecting a  specific  collection of events in a new design.  A designer may, for example,  consider  the way  occupants interact with light switches on arriving inside a building in summer mornings to be  of crucial importance to a design purpose. In this case, the relevant contextual factor, namely, summer mornings, should be  explicitly present in a design to record occupant interactions. The knowledge underlying context-aware design-specific data  helps augment an existing design for better customization with respect to  a new building design context. To acquire context-aware design-specific data, An IVE provides a mechanism for conducting stated choice experiments for acquiring context-aware design-specific data, e.g., occupant’s use of artificial lighting on a clear summer day. 
\subsection{Incorporate the identified contextual knowledge into existing design models for improved performance}
 A  computational framework \cite{chana} based on generative adversarial networks (GANs) \cite{goodfellow2014generative} is yused to reduce the discrepancy between design-time predictions of system performance  and  that obtained during actual system operation by fusing knowledge about human interactions with  contextual and design-specific factors of new systems under design with that of existing instances. IVEs are used as platforms for performing stated choice  behavioral experiments  to acquire human-interaction data. The use of Generative Adversarial Networks (GANs) enables  learning to generate mixture models that  appropriately fuse existing design models with knowledge of human interactions obtained from the  stated choice experiments on the IVE to create augmented design models with enhanced predictive power. Performance targets and design criteria, specified by a designer, are used as guides to achieve appropriate combination. Performance targets can be obtained from benchmarks in literature  or can be based on standards mandated by the Department of Energy or similar agencies \cite{attia2012development}. They can provide benchmarks for energy savings or comfort levels and similar other metrics. 
A GAN consists of two neural networks: a generator and a discriminator. In our framework, the generator  is provided as input two data sets:  a data set sampled from an existing design model (called existing dataset) and that obtained from stated choice human experiments from the  IVE  (called the IVE dataset). It  outputs a mixture  probability distribution (called augmented model; see Fig \ref{comp_result2}). The performance predictions obtained from the augmented should be as close as possible to the performance targets provided. The discriminator  tries to discriminate between the performance predictions obtained from the augmented model produced  by the generator and the performance targets. The GAN is trained through a minimax game between the generator and the discriminator, during which the generator attempts to generate an  augmented model whose predictions are in conformance with the  performance targets. The discriminator attempts to determine if the predictions of the augmented model conforms with the performance targets.  The training continues until a defined stopping criterion (maximum number of epochs or discrepancy measured between the predictions of the augmented model and the targets is below a threshold) is reached. Once training converges, the distribution of the resulting generator is   the augmented model.
The generator $G$ using the existing dataset $x$ and the IVE dataset $z$ as input learns to generate an appropriate mixture of the two to meet the performance target. To learn a generator distribution $\pi_g$, the generator learns a mapping function from product space the existing dataset (with empirical probability distribution $\pi_x$) and the IVE dataset (with the empirical probability distribution $pi_z$) to  the generator data space  $G(x,z; \theta_g)$ (where $\theta_g$ constitutes the parameters of $G$). The discriminator $D(G(x,z; \theta_g); \theta_d)$  (where $\theta_d$ represents the parameters of the discriminator) will output a single scalar that determines whether the performance targets are met or not. We  partially adapt the concept of conditional GANs \cite{mirza2014conditional} by using information of input features of the generator as additional inputs to the discriminator model.  
\subsection{A closed loop framework for improving building design that uses feedback from physical experiments to identify high-fidelity contextual factors to improve future design models}
Contextual factors (e.g., occupancy status, short term leaving, and work area illuminance, in case of light switching predictions) have a different level of causal impact on predictions. Thus, it is important to consider feedback from physical experiments to determine the relative importance, in terms of causality \cite{alma,hernan}, of such factors. In addition, the set of contextual variables considered may be incomplete or inappropriate. Feedback from physical experiments can guide the choice of high-fidelity contextual variables. To generate feedback from physical experiments, we  use causal analysis. To the best of our knowledge,   not much research has been performed in  applying Causal Analysis methods to understand the influence of context factors on the performance of a CPHS under design.    Causal Analysis techniques enable us to identify the contextual factors that influence a design and  it subsequently allows us to enhance design models that  better forecast the performance of a CPHS under design and also to better understand human interactions. The main objective is to conduct Causal Analysis of human interactions using data collected from Stated Choice Experiments based on the augmented model in a physical experiment facility.  In causal inference, we need to formally represent our assumptions about causal relationship within data. This is achieved through Graphical Models \cite{alma,scheines1997introduction,pearl2009causality,pearl2016causal}.  Directed Acyclic Graphs (DAGS) can model probability distributions underlying a data set and provide the foundations of  causal graphical models.  Causal assumptions are expressed using an iterative procedure following three steps \cite{nabijiang2019you}. We  use the contextual variables from the augmented model and construct our pilot causal graph based on that model. In the second step, since a causal graph models testable implications, we  can  test our assumptions to some extent using graphical criteria on the data collected from the stated choice experiments in a physical experiment facility. In the final step, we  evaluate the pilot model. We  modify our graph according to the results obtained by testing the model against data collected from Stated Choice Experiments in a physical experiment facility. This modification involves introducing new context variables, removing redundant variables, or modifying the relationship between variables by adding or eliminating nodes and edges. After proper adjustments, we  finalize the causal model for further causal inference procedure. We can test the pilot causal model stated above by applying the criterion of d-separation \cite{alma}. This enables checking if the model conforms to the data. The basic idea of d-separation is to identify the common causes and common effects that are the sources of confounding and selection bias, and if present, to block (or separate) their association flow in the path so that we can estimate the actual causal effect. 
Average causal effects can be determined from the  difference of the values of counter-factual outcomes \cite{rubin1974estimating,robins1994estimation}. For counter-factual outcomes, we  use the contextual variables (for example occupancy status in case of predicting light switching) in the augmented model as treatments. To calculate the causal effect, we  use inverse probability (IP) weighting method \cite{robins1994estimation}. The purpose of using Inverse probability weight is to break the association between the covariates and treatment. To determine the causal effect, we  create a logistic regression model, and then use its estimated probability values as weights for further analysis.  We  determine each variable's causal effect on outcome, e.g., the effect of occupancy status on light switching. To determine the confounding variables, we  pair every independent variable with the outcome variable, separate non-causal paths between them and select the confounding variables using graph surgery techniques \cite{pearl2014probabilistic}. 
Results obtained from the causal analysis described above (in particular the final causal graph and the causal effect analysis) are sent back as feedback to the IVE and will be used to design the IVE experiments  in the next iteration. The iterations  continue until the difference between the pilot causal model (from the augmented design model) and the final causal model (after causal validation on the stated choice experiment data from physical experiment facility) is minimal or the designer is satisfied with the performance of the augmented model. 
\subsection{Performance Targets}
During design, designers take into account and trade-off multiple factors (e.g., code compliance, comfort, cost, energy, function, operation, occupancy characteristic, and sustainability) to establish  the goals and objectives of a system. In fact, constructing performance targets depends on design goals and objectives of the system. 
An alternative to constructing performance targets is that designers may gather information of important factors and  estimate system performance with respect to its goals and objectives. Designers or researchers correlate such factors using statistical analysis such as regression models and use these models as performance targets. For instance, to create a performance target of lighting usages, designers or researchers collect information regarding to building lighting designs, which may include historical human-building lighting interaction in existing buildings, that have similar characteristics as the building under design.
\section{A Case Study of CPHS: Energy-Efficient Building Design}
A small case study describing the design of the lighting system of an energy-efficient building using the CPHS framework is presented below \footnote{The case study received approval from the Institutional Review Board at Louisiana State University}.
\subsection{IVE Design}
The IVE has been  developed using the STED model \cite{saeidi2018spatial}.  
The STED model consists of two structures, an event structure and a spatial-temporal structure. The event structure organizes critical events of a building lifecycle into a hierarchical structure according to  research needs. For example, using lighting as a case, a year can be divided into four seasons; each season may be modeled using three scenarios, no artificial lighting, possible artificial lighting, and artificial lighting (i.e., based on outdoor daylighting and the ASHREA standards 90.1 to determine if indoor artificial lighting is necessary). A scenario refers to a time point from 00:00 to 24:00. At each time point, an office space configuration may be modeled according to research needs. For example, we may consider two factors, ceiling lights and blinds. Regardless of outdoor daylighting conditions, there are four indoor configurations based on the on/off status of the ceiling lights and the up/down status of the blinds.  On a bright spring morning, when indoor artificial lighting may not be needed, the office space configuration may have the ceiling lights off and blinds up, or ceiling lights on and blinds up, or any other combinations. Since the combination of daylighting scenarios and configurations result in different levels of  visual comfort inside the space, it is expected that a subject will operate the lighting switch differently.
In this example, outdoor lighting conditions, occupancy, and status of lighting and blinds are contextual factors with respect to the Hunt’s model which only considers work area illuminance level. If for a new design, some of those contextual factors may potentially be significant, a method to identify them and include them in a design evaluation is critical. 

The setup of the actual IVE experiments \cite{chana} were guided by  critical events in the data acquired  from the physical environment (e.g., arrival at the office, intermediate leaving, coming back from intermediate leaving, and departure). Each event corresponds to a tuple of  values for the contextual factor variables (e.g., indoor and outdoor illuminance, intermediate leaving status, and occupancy status) in new-design buildings. Occupants were presented with the contextual factors in  event based experiments. The occupant’s interactions with the light switch were recorded. For example, the occupant turned on  the light  when both indoor and outdoor were dark.
\subsection{Data from IVE Experiments}
A total of 180 data points \cite{chana} relating to occupant preferences (lighting) and values of contextual factor variables (indoor and outdoor illuminance, intermediate leaving status, and occupancy status) were acquired from the IVE experiments; 36 initial events before arrival at the office, 36 events of arrival at the office, 18 events of intermediate short leave, 18 events of returning from the intermediate short leave, 18 events of intermediate long leave, 18 events of returning from the intermediate long leave, and 36 events of departure. 
\subsection{Combining Existing Design Model with Data from IVE experiments}
We used the Hunt's model \cite{hunt1980predicting} as the  existing design model based on historical data. The GAN-based computational framework  of CPHS combined this model with the knowledge obtained from the data from IVE experiments with a performance target as a guide to obtain an augmented model. For the performance target, we use the probabilities of switching on as provided by a probit model described in \cite{da2013occupants}. 
\begin{figure}[H]
\centering
    \begin{adjustbox}{width=\columnwidth}
	\begin{tikzpicture}
	\begin{axis}[
	xmin=200, xmax=700, ymin=-0.05,ymax=1.05,
	legend style={at={(0.6,0.15)},anchor=south,draw=none,fill=none,column sep=10pt, nodes={scale=0.8, transform shape}},
	legend cell align={left},
	xlabel = \textbf{Work area illuminance (Lux)},
	ylabel = \textbf{Probability of switching on}]
	
	\addplot +[only marks,mark=square*,mark size=2pt] table[col sep=comma, x = W1, y = Pe1]{data1.txt};
	\addlegendentry{Performance Target};
	
	\addplot +[only marks,mark=triangle*,mark size=2pt] table[col sep=comma, x = W1, y = Au1]{data1.txt};
	\addlegendentry{Augmented Design Model};
	
	\addplot +[only marks,mark=otimes*,mark size=2pt] table[col sep=comma, x = W1, y = Ex1]{data1.txt};
	\addlegendentry{Existing Design Model};
	
	\addplot +[only marks,mark=diamond*,mark size=2pt] table[col sep=comma, x = W1, y = Sy1]{data1.txt};
	\addlegendentry{Data from IVE Experiment};
	\end{axis}
	\end{tikzpicture}
	\end{adjustbox}
\caption{Result of Experiments. \cite{chana}}
\label{fig2}
\end{figure}
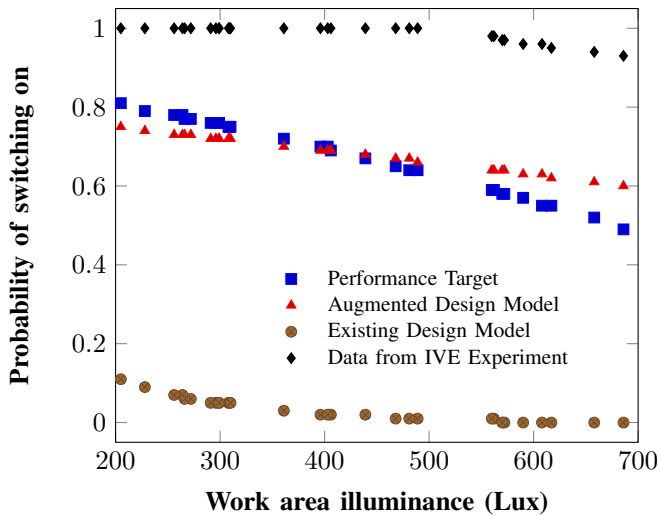
\subsection{Experimental Results}
Fig \ref{fig2} \cite{chana} shows the experimental results from our case study. It can be seen that the predictions of the augmented design model are much closer to the performance target than that of the existing design model (Hunt's model). 

\section{Related Work}
Generative Adversarial Networks (GANs) were introduced in \cite{goodfellow2014generative}. GANs have applications in  various areas especially image optimization and generation \cite{ledig2017photo,icadl,radford2015unsupervised}. The main purpose of our research is to create augmented design models whose predictions are close to the performance targets from the mixture of existing predictive models supporting the design process and the knowledge of  human interactions with new designs in response to contextual factors.  GANs can be used for generating mixture models to serve the purposes of our research in the computational framework \cite{chana}. Deep learning methods \cite{lecun2015deep,collier,collier2,Saikat,DeepSAT,manohar,boyda2017deploying} can be used to enhance CPHS frameworks by extracting information about physical contexts such as terrain, outside environmental conditions, etc. 

Causal inference  has been used in the social and behavioral sciences for developing and updating theories as well as public policies \cite{holland1986statistics,morgan2015counterfactuals}. In \cite{nabijiang2019you}, we used causal reasoning for understanding drivers’ route choice behavior in context-aware transportation systems. In the present framework, we  use causal models to assess the impact of contextual factors on system performance. Based on this assessment, feedback is generated that will be used in the next iteration to develop design models with improved predictive powers. There has been some work on understanding user preference behavior in the context of human-robot interaction \cite{zahedi2019towards}.  In previous work \cite{qun}, we used knowledge distillation to improve existing route choice models in transportation systems by incorporating context factors. 

To acquire data specific to a design, the CPHS framework provides an approach to study and record human-system interaction  during  designs through  a combination of immersive virtual environments (IVEs) and machine learning.  Stated choice IVE experiments are appropriate for acquiring data human-infrastructure interactions at design time due to several factors.  For instance, IVEs enable users to control confounding and isolating variables of interest, to be immersed in their settings, and to continually maintain variables of interest during conducting experiments \cite{maldovan2006framework}.
Heydarian et al. \cite{heydarian2015immersive,heydarian2017towards}  and Saeidi et al. \cite{saeidi2015measuring}  used IVEs to study occupant behaviors related to lighting and shade usages in buildings.
Chokwitthaya et al. \cite{chokwitthaya2019combining,chokwitthaya2019machine} explored using machine learning to improve the prediction of artificial lighting usage during design by combining an existing model with data from IVE experiments.  The examples show that IVEs have the potential of replicating experiences in physical environments, acquiring design specific data, and improving the prediction of human interactions during design. 
\section{Conclusions and Discussion}
Even though the results of the study are promising, IVE-based SCE’s have their inherent limitations. Due to limitations of IVE technologies, it is difficult to continuously collect data on interaction of humans with cyber-physical systems. Our current work is exploring the use of Shannon-Nyquist sampling techniques to determine the optimal rate of sampling for effective inference. Due to the expensive nature of the IVE SCEs, we could only explore a limited set of scenarios. While it showed the potential of the framework, there is scope for computational bias and overfitting. In \cite{qun}, we have attempted to reduce the size of the model using knowledge distillation to counter overfitting. However, more scenarios need to be considered. In addition, bias-variance-cost trade-offs need to be investigated.  Also, due to the time and cost limitations, only a small sample size of participants (180) was considered.  To this end, low shot and unsupervised  \cite{unsupervised,ravi2016optimization} learning techniques need to be considered to learn from a small dataset. 

\bibliographystyle{IEEEtran}
\bibliography{IJCNN_BIB,CogSci_Template}

\end{document}